\documentclass[12pt,twocolumn]{article}

\usepackage{booktabs}
\usepackage[utf8]{inputenc}
\usepackage{gvsetsTitle}
\usepackage[none]{hyphenat}
\usepackage{enumerate}
\usepackage{listings}
\usepackage{times}
\usepackage[paperwidth=8.5in, paperheight=11in, margin=0.5in]{geometry}
\usepackage{indentfirst}
\usepackage{fancyhdr}
\usepackage{lastpage}
\fancypagestyle{plain}{ %
  \fancyhf{}
}
\renewcommand{\footnoterule}{%
  \kern -3pt
  \hrule width \columnwidth
  \kern 2pt
}
\usepackage{authblk}
\usepackage{graphicx}
\usepackage{float}
\usepackage{hyperref}
\usepackage{titlesec}
\titlelabel{\thetitle.\quad}
\usepackage{amsmath,amssymb,amsfonts}
\usepackage{algorithmic}
\usepackage{algorithm}
\usepackage{array}
\usepackage{subcaption}
\usepackage{textcomp}
\usepackage{stfloats}
\usepackage{url}
\usepackage{verbatim}
\usepackage{cite}
\usepackage{pifont}
\usepackage{multirow}
\usepackage{balance}
\usepackage{mathrsfs}

\session{Modeling, Simulation, Prototyping, and Validation} 

\title{Off-Road Autonomy Validation Using Scalable Digital Twin Simulations Within High-Performance Computing Clusters}

\author[1]{Tanmay Samak} 
\author[1]{Chinmay Samak}
\author[2]{Joey Binz}
\author[3]{Jonathon Smereka}
\author[3]{Mark Brudnak}
\author[3]{David Gorsich}
\author[2]{Feng Luo}
\author[1]{Venkat Krovi}
\affil[1]{Department of Automotive Engineering, CU-ICAR, Greenville, SC} 
\affil[2]{School of Computing, Clemson University, Clemson, SC}
\affil[3]{U.S. Army DEVCOM Ground Vehicle Systems Center, Detroit, MI}

\begin{document}
\twocolumn[  
    \begin{@twocolumnfalse}
    \maketitle
    \thispagestyle{fancy}
    
    \vspace{0.25em}
    \begin{abstract}
      Off-road autonomy validation presents unique challenges due to the unpredictable and dynamic nature of off-road environments. Variability analyses, by sequentially sweeping across the parameter space, struggle to comprehensively assess the performance of off-road autonomous systems within the imposed time constraints. This paper proposes leveraging scalable digital twin simulations within high-performance computing (HPC) clusters to address this challenge. By harnessing the computational power of HPC clusters, our approach aims to provide a scalable and efficient means to validate off-road autonomy algorithms, enabling rapid iteration and testing of autonomy algorithms under various conditions. We demonstrate the effectiveness of our framework through performance evaluations of the HPC cluster in terms of simulation parallelization and present the systematic variability analysis of a candidate off-road autonomy algorithm to identify potential vulnerabilities in the autonomy stack's perception, planning and control modules.\\
    \end{abstract}
    
    \vspace{0.25em}
    \begin{adjustwidth}{0.5in}{0.5in}
    \noindent\small\textbf{Citation:} T. Samak, C. Samak, J. Binz, J. Smereka, M. Brudnak, D. Gorsich, F. Luo and V. Krovi, ``Off-Road Autonomy Validation Using Scalable Digital Twin Simulations Within High-Performance Computing Clusters,'' In \textit{\proceedings}, NDIA, Novi, MI, \gvsetsdates, \gvsetsyear.
    \end{adjustwidth}
    \vspace{1em}
    \end{@twocolumnfalse}
]

\vspace{1em}




{
    \let\thefootnote\relax\footnotetext{
    This work was supported by the Virtual Prototyping of Autonomy Enabled Ground Systems (VIPR-GS), a U.S. Army Center of Excellence for modeling and simulation of ground vehicles, under Cooperative Agreement W56HZV-21-2-0001 with the U.S. Army DEVCOM Ground Vehicle Systems Center (GVSC).
    
    DISTRIBUTION STATEMENT A. Approved for public release; distribution is unlimited. OPSEC \#8451.
    }
}


\section{Introduction}
\label{Section: Introduction}

Modeling and simulation of autonomous vehicles \cite{Schoner2018} plays a crucial role in achieving enterprise-scale realization that aligns with technical, business and regulatory requirements. Contemporary trends in digital lifecycle treatment have proven beneficial to support simulation-based design (SBD) as well as verification and validation (V\&V) of increasingly complex systems and system-of-systems. Most enterprise-scale V\&V strategies \cite{7795548, Pathrose2022} adopt the ISO26262 V-model \cite{ISO26262} or one of its derived forms \cite{AutoDRIVEMechatronics} to lay down the functional requirements, which are then decomposed into system designs and verified. This is followed by the elucidation of sub-system requirements and designs all the way down to individual component requirements and designs, with a verification step at each stage. The components are then developed and tested at the unit level, following which, they are integrated into sub-systems, systems, and potentially system-of-systems. Each integration stage is followed by a validation stage, with the ultimate objective of rolling out a reliable product.

The first roadblock in terms of digitizing V\&V strategies is the development of appropriate fidelity simulation models capable of capturing the intricate real-world physics and graphics (real2sim), while enabling real-time interactivity for decision-making, has remained a challenge. Autonomy-oriented digital twins \cite{samak2024validation, samak2024autonomy}, as opposed to conventional simulations, must equally prioritize back-end physics and front-end graphics, which is crucial for the realistic simulation of vehicle dynamics, sensor characteristics, and environmental physics. Additionally, the interconnect between vehicles, sensors, actuators and the environment, along with peer vehicles and infrastructure in a scene must be appropriately modeled. Most importantly, however, these simulations should allow real-time interfacing with software development framework(s) to support reliable verification and validation of autonomy algorithms.

To this end, recent advances in artificial intelligence (AI) based tools and workflows, such as online deep-learning algorithms leveraging live-streaming data sources, offer the tantalizing potential for real-time system identification and adaptive modeling to simulate vehicle(s), environment(s), as well as their interactions. This transition from static/fixed-parameter ``virtual prototypes'' to dynamic/adaptable ``digital twins'' not only improves simulation fidelity and real-time factor but can also support the development of online adaption/augmentation techniques that can help bridge the gap between simulation and reality (sim2real) \cite{AutoDRIVESim2Real2023}.

\begin{figure}[t]
    \centering
    \includegraphics[width=\linewidth]{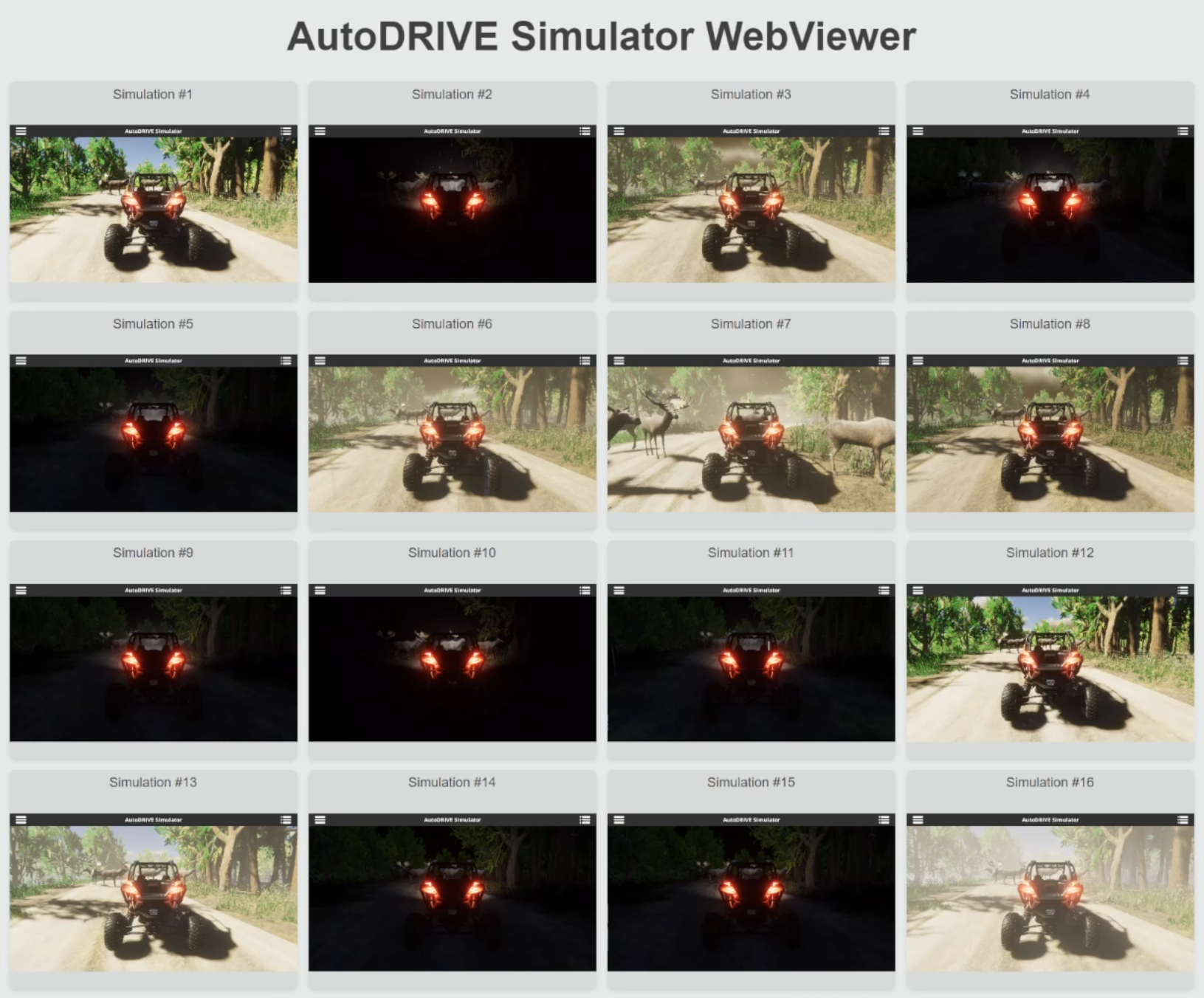}
    \caption{Interactive web application visualizing the variability testing of autonomous RZR digital twin across different weather conditions and times of the day using parallel simulations in the cloud.}
    \label{fig1}
\end{figure}

However, digital twins can only solve half the problem. Validation of autonomous vehicles, especially in off-road conditions, requires robust testing across a wide variety of conditions and edge cases. Constructing a test matrix tailored to the given specifications and subsequently executing these tests sequentially, either manually \cite{RBC2021} or using an automated workflow \cite{VVChallenges}, can take several days.

Fortunately, contemporary technologies like cloud computing and parallel processing can help alleviate this pain point. This leads to the emergence of utilizing modeling and simulation as a service (MSaaS) \cite{6721436, 7988851}, which significantly expedites the autonomy validation process. MSaaS involves orchestrating simulation resources in the cloud, which are managed by a service provider, ensuring automatic, elastic, and dependable provisioning of computing resources based on user demand. Such characteristics of MSaaS facilitate efficient resource management and utilization, leading to a significant reduction in testing time and cost. Specifically within the realm of autonomous vehicles, MSaaS brings forth several features that significantly compress the timeframe between development and deployment:
\begin{itemize}
    \item\textbf{Scalability:} Empowering the execution of hundreds of tests within a single cycle, ultimately facilitating the accumulation of millions of virtual test miles. This aids in discerning isolated outcomes, delineating performance boundaries, and identifying system tolerances.
    \item\textbf{Parameterization:} Enabling the execution of simulations across a spectrum of environmental parameters (e.g., time of day, weather, traffic, road conditions, pedestrians, etc.) and vehicle configurations (e.g., vehicle dynamics, sensor placement, network latency, etc.), a practice often referred to as a parameter sweep.
    \item\textbf{Continuous Testing and Integration:} Facilitating seamless regression testing, agile workflows, version control, data tagging, and more with each iteration of changes in vehicle software, sensors, or infrastructure.
\end{itemize}

This study aims to elucidate the importance of MSaaS in the development and validation of autonomous vehicle software, exemplifying its effectiveness through a case study. Particularly, this work presents a modular and open-source framework for MSaaS and demonstrates its effectiveness through the systematic V\&V of an autonomous light tactical vehicle (LTV) operating in an off-road environment. Following are the key contributions of this work:
\begin{itemize}
    \item Developing a high-fidelity and photorealistic digital twin simulation framework for off-road autonomous vehicles.
    \item Setting up a cloud infrastructure for on-demand elastic orchestration of containerized simulation instances as well as an interactive web-viewer application using HPC resources.
    \item Demonstrating the end-to-end workflow of validating a candidate off-road autonomy algorithm from test-case definition and generation to autonomy V\&V and computational analysis.
\end{itemize}

The remainder of this paper is organized as follows. Section \ref{Section: Related Work} summarizes state-of-the-art literature pertaining to simulation frameworks and their containerized orchestration in HPC settings. Section \ref{Digital Twin Framework} elucidates the high-fidelity simulation of vehicle, sensors and environment models. Section \ref{Section: HPC Deployment Framework} delves into containerization, configuration and orchestration of parallel simulations in the cloud. Section \ref{Section: Results and Discussion} presents performance evaluations of the HPC cluster in terms of simulation parallelization and systematic variability analysis of a candidate off-road autonomy algorithm. Finally, Section \ref{Section: Conclusion} summarizes the work and points towards potential research directions.


\section{Related Work}
\label{Section: Related Work}

We summarize the existing literature in two distinct sections, although they do have an overlap. Section \ref{Section: Simulation Frameworks} describes the state-of-the-art simulation frameworks employed for SBD as well as V\&V of autonomous ground vehicles. Section \ref{Section: HPC Frameworks} then delves into the prior work pertaining to HPC deployments as well as simulator containerization and orchestration.

\subsection{Simulation Frameworks}
\label{Section: Simulation Frameworks}

Automotive industry has employed simulators like Ansys Automotive \cite{AnsysAutomotive2021} and Adams Car \cite{AdamsCar2021} to simulate vehicle dynamics at different levels, thereby accelerating the development of its end-products. Since the past few years, however, owing to the increasing popularity of advanced driver-assistance systems (ADAS) and autonomous driving (AD), most of the traditional automotive simulators, such as Ansys Autonomy \cite{AnsysAutonomy2021}, CarSim \cite{CarSim2021} and CarMaker \cite{CarMaker2021}, have started releasing vehicular autonomy features in their updated versions.

Apart from these, several commercial simulators specifically target autonomous driving. These include NVIDIA's Drive Constellation \cite{DRIVEConstellation2021}, Cognata \cite{Cognata2021}, rFpro \cite{rFpro2021}, dSPACE \cite{dSPACE2021} and PreScan \cite{PreScan2021}, to name a few. In the recent past, several research projects have also tried adopting computer games like GTA V \cite{Richter2016, Richter2017, Johnson-Roberson2017} in order to virtually simulate self-driving cars, but they were quickly shut down.

Lastly, the open-source community has also contributed several simulators for such applications. Gazebo \cite{Gazebo2004} is a generic robotics simulator natively adopted by Robot Operating System (ROS) \cite{ROS1}. TORCS \cite{TORCS2021} is probably one of the earliest simulation tools to specifically target manual and autonomous racing problems. More recent examples include CARLA \cite{CARLA2017}, AirSim \cite{AirSim2018} and Deepdrive \cite{Deepdrive2021} developed using the Unreal \cite{Unreal2021} game engine along with Apollo GameSim \cite{ApolloGameSim2021}, LGSVL Simulator \cite{LGSVLSimulator2020} and AWSIM \cite{AWSIM2023} developed using the Unity \cite{Unity2021} game engine.

The aforementioned simulators pose three key limitations, which are addressed by this work:

\begin{itemize}
    \item Firstly, some simulation tools prioritize graphical photorealism at the expense of physical accuracy, while others prioritize physical fidelity over graphical realism. In contrast, AutoDRIVE Simulator \cite{AutoDRIVESimulator, AutoDRIVESimulatorReport} achieves a harmonious equilibrium between physics and graphics, offering a variety of configurations to suit diverse computational capabilities.
    \item Secondly, the perception as well as dynamics characteristics of off-road autonomous vehicles and their operating environments differ significantly from traditional ones. Existing simulators primarily target the on-road autonomy segment and those that target the off-road operational design domain (ODD) focus primarily on vehicle dynamics and terramechanics with limited attention to rendering and photorealism. Consequently, transitioning autonomy algorithms from such simulators to the field necessitates considerable additional effort to re-calibrate (or potentially re-design) the autonomy algorithms.
    \item Thirdly, existing simulators may lack precise representations of real-world vehicles or environments, rendering them unsuitable for ``digital twinning'' applications.
\end{itemize}

\subsection{HPC Frameworks}
\label{Section: HPC Frameworks}

The increasing prevalence of the software as a service (SaaS) paradigm \cite{tsai2014software} and the wide adoption of services such as Microsoft Azure, Google Cloud, and Amazon Web Services (AWS) \cite{gupta2021review} allows the utilization of immense compute power by virtually any pipeline or workflow adopted to do so. Existing works have sought to utilize HPC resources for modeling and simulation in a variety of ways, many of which have a large potential to save time and money when compared to non-HPC workflows accomplishing similar tasks \cite{science2019modelling, lyu2020evaluation, franchi2022webots}.

Many approaches utilize container orchestration frameworks such as Kubernetes \cite{kubernetes2019kubernetes}, which is widely used due to its open-source nature, self-healing capability, scalability, rich ecosystem of baked-in tools, and active community. However, it can infamously have a steep learning curve due to its complex nature. Container orchestration frameworks all need robust container runtime solutions so that programs can run with proper dependencies on distributed computing. Docker \cite{merkel2014docker} and Singularity \cite{singularity2017} are two of the most widely used solutions for containerization.

A common theme among previous works is simulating multiple sub-systems of a larger system by isolating each subsystem into a separate Kubernetes pod. A project by Fogli et al. leveraged this approach to validate distributed industrial production systems \cite{fogli2023chaos}. Another project by Rehman et al. used this approach to co-simulate systems for a corporate electric vehicle fleet \cite{rehman2019cloud}. Other approaches have simulated and tested the performance of Kubernetes clusters themselves \cite{khan2021perfsim}.

Our work seeks to remedy the existing limitations and establish originality in two key aspects:
\begin{itemize}
    \item Firstly, previous works have been very specialized. We aim to present a pipeline for scalable autonomous driving simulations built on widely used platforms, usable for a large variety of purposes.
    \item Secondly, integration of existing pipelines into simulators not custom-built for cloud computing infrastructures is lacking. Our framework is capable of deploying non-cloud-native applications and seeks to be easily adaptable to any existing ground vehicle simulator that extends an application programming interface (API).
\end{itemize}


\section{Digital Twin Framework}
\label{Digital Twin Framework}

The automotive industry has long practiced a gradual transition from virtual, to hybrid, to physical prototyping within an X-in-the-loop (XIL; X = model, software, processor, hardware, vehicle) framework. More recently, digital twins have emerged as potentially viable tools to improve simulation fidelity and to develop adaption/augmentation techniques that can help bridge the sim2real gap. In the following sections, we delve into the development of a high-fidelity digital twin of an autonomous LTV, namely Polaris RZR PRO R 4 ULTIMATE, and one of its operating environments using AutoDRIVE Ecosystem \cite{AutoDRIVEEcosystem, AutoDRIVEReport}.

\subsection{Vehicle}
\label{Vehicle}

\begin{figure*}[t]
    \centering
    \includegraphics[width=\linewidth]{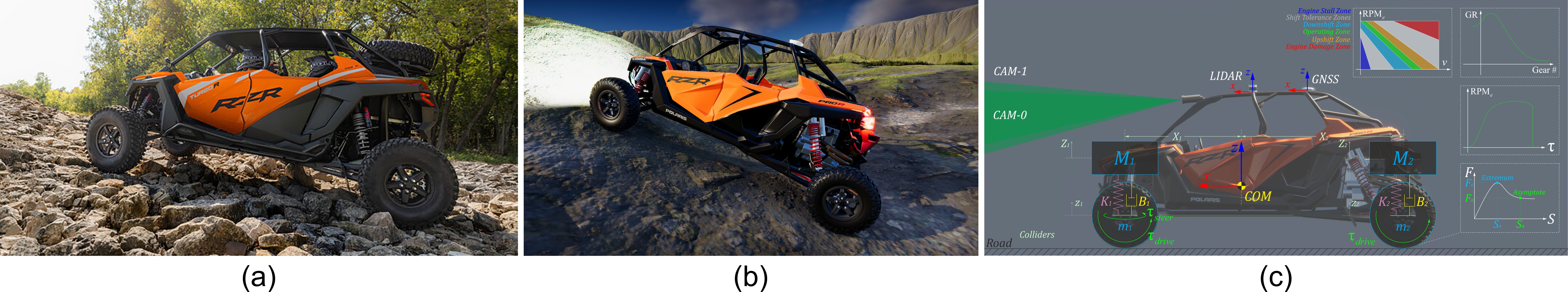}
    \caption{Autonomy-oriented digital twin of the Polaris RZR PRO R 4 ULTIMATE: (a) physical RZR in the real world, (b) digital twin of RZR in AutoDRIVE Simulator, and (c) simplified representation of the vehicle dynamics and sensor simulation models.}
    \label{fig2}
\end{figure*}

The vehicle (refer Fig. \ref{fig2}) is conjunctly modeled using sprung-mass ${^iM}$ and rigid-body representations. Here, the total mass $M=\sum{^iM}$, center of mass, $X_\text{COM} = \frac{\sum{{^iM}*{^iX}}}{\sum{^iM}}$ and moment of inertia $I_\text{COM} = \sum{{^iM}*{^iX^2}}$, serve as the linkage between these two representations, where ${^iX}$ represents the coordinates of the sprung masses. Each vehicle's wheels are also modeled as rigid bodies with mass $m$, experiencing gravitational and suspension forces: ${^im} * {^i{\ddot{z}}} + {^iB} * ({^i{\dot{z}}}-{^i{\dot{Z}}}) + {^iK} * ({^i{z}}-{^i{Z}})$.

\subsubsection{Suspension Dynamics}
\label{Suspension Dynamics}

The stiffness ${^iK} = {^iM} * {^i\omega_n}^2$ and damping $^iB = 2 * ^i\zeta * \sqrt{{^iK} * {^iM}}$ coefficients of the suspension system are computed based on the sprung mass ${^iM}$, natural frequency ${^i\omega_n}$, and damping ratio ${^i\zeta}$ parameters. The point of suspension force application ${^iZ_F}$ is calculated based on the suspension geometry:
${^iZ_F} = {^iZ_\text{COM}} - {^iZ_w} + {^ir_w} - {^iZ_f}$, where $^iZ_\text{COM}$ denotes the Z-component of vehicle's center of mass, $^iZ_w$ is the Z-component of the relative transformation between each wheel and the vehicle frame ($^VT_{w_i}$), $^ir_w$ is the wheel radius, and $^iZ_f$ is the force offset determined by the suspension geometry. Lastly, the suspension displacement $^iZ_s$ at any given moment can be computed as ${^iZ_s} = \frac{{^iM} * g}{{^iZ_0} * {^iK}}$, where $g$ represents the acceleration due to gravity, and $^iZ_0$ is the suspension's equilibrium point. Additionally, the vehicle model also has a provision to include anti-roll bars, which apply a force on the left ${^LF_r} = K_r * {^RZ} - {^LZ}$ and right ${R^F_r} = K_r * {^LZ} - {^RZ}$ wheels as long as they are grounded at the contact point $Z_c$. This force is directly proportional to the stiffness of the anti-roll bar, $K_r$. The left and right wheel travels are given by ${^LZ} = \frac{-{^LZ_c} - {^Lr_w}}{^LZ_s}$ and ${^RZ} = \frac{-{^RZ_c} - {^Rr_w}}{^RZ_s}$.

\subsubsection{Powertrain Dynamics}
\label{Powertrain Dynamics}

The powertrain comprises an engine, transmission and differential. The engine is modeled based on its torque-speed characteristics. The engine RPM is updated smoothly based on its current value $RPM_e$, the idle speed $RPM_i$, average wheel speed $RPM_w$, final drive ratio $FDR$, current gear ratio $GR$, and the vehicle velocity $v$. The update can be expressed as $RPM_e := \left[RPM_i + \left(|RPM_w| * FDR * GR\right)\right]_{(RPM_e,v)}$ where, $[\mathscr{F}]_x$ denotes evaluation of $\mathscr{F}$ at $x$.

The total torque generated by the powertrain is computed as $\tau_{\text{total}} = \left[\tau_e\right]_{RPM_e} * \left[GR\right]_{G_\#} * FDR * \tau * \mathscr{A}$. Here, $\tau_e$ is the engine torque, $\tau$ is the throttle input, and $\mathscr{A}$ is a non-linear smoothing operator which increases the vehicle acceleration based on the throttle input.

The automatic transmission decides to upshift/downshift the gears based on the transmission map of a given vehicle. This keeps the engine RPM in a good operating range for a given speed: $RPM_e = \frac{{v_{\text{MPH}} * 5280 * 12}}{{60 * 2 * \pi * R_{\text{tire}}}} * FDR * GR$. It is to be noted that while shifting the gears, the total torque produced by the powertrain is set to zero to simulate the clutch disengagement. It is also noteworthy that the auto-transmission is put in neutral gear once the vehicle is in standstill condition and parking gear if handbrakes are engaged in standstill condition. Additionally, switching between drive and reverse gears requires that the vehicle first be in the neutral gear to allow this transition.

The total torque $\tau_\text{total}$ from the drivetrain is divided to the wheels based on the drive configuration of the vehicle:
$
\tau_{\text{out}} = \begin{cases}
\frac{\tau_{\text{total}}}{2} & \text{if FWD/RWD} \\
\frac{\tau_{\text{total}}}{4} & \text{if AWD}
\end{cases}
$.
The torque transmitted to wheels $\tau_w$ is modeled by dividing the output torque $\tau_\text{out}$ to the left and right wheels based on the steering input. The left wheel receives a torque amounting to $^{L}\tau_{w} = \tau_{\text{out}} * (1 - \tau_{\text{drop}} * |\delta^{-}|)$, while the right wheel receives a torque equivalent to $^{R}\tau_{w} = \tau_{\text{out}} * (1 - \tau_{\text{drop}} * |\delta^{+}|)$. Here, $\tau_\text{drop}$ is the torque-drop at differential and $\delta^{\pm}$ indicates positive and negative steering angles, respectively. The value of $(\tau_{\text{drop}} * |\delta^{\pm}|)$ is clamped between $[0,0.9]$.

\subsubsection{Steering Dynamics}
\label{Steering Dynamics}

The steering mechanism operates by employing a steering actuator, which applies a torque $\tau_{\text{steer}}$ to achieve the desired steering angle $\delta$ with a smooth steering rate $\dot{\delta}$, without exceeding the steering limits $\pm \delta_\text{lim}$. The rate at which the vehicle steers is governed by its speed $v$ and steering sensitivity $\kappa_\delta$, and is represented as $\dot{\delta} = \kappa_\delta + \kappa_v * \frac{v}{v_\text{max}}$. Here, $\kappa_v$ is the speed-dependency factor of the steering mechanism. Finally, the individual angle for left $\delta_l$ and right $\delta_r$ wheels are governed by the Ackermann steering geometry, considering the wheelbase $l$ and track width $w$ of the vehicle:
$
\left\{
\begin{matrix} 
\delta_l = \textup{tan}^{-1}\left(\frac{2*l*\textup{tan}(\delta)}{2*l+w*\textup{tan}(\delta)}\right) \\ 
\delta_r = \textup{tan}^{-1}\left(\frac{2*l*\textup{tan}(\delta)}{2*l-w*\textup{tan}(\delta)}\right) 
\end{matrix}
\right.
$.

\subsubsection{Brake Dynamics}
\label{Brake Dynamics}

The braking torque is modeled as ${^i\tau_\text{brake}} = \frac{{^iM}*v^2}{2*D_\text{brake}}*R_b$ where $R_b$ is the brake disk radius and $D_\text{brake}$ is the braking distance at 60 MPH, which can be obtained from physical vehicle tests. This braking torque is applied to the wheels based on the type of brake input: for combi-brakes, this torque is applied to all the wheels, and for handbrakes, it is applied to the rear wheels only.

\subsubsection{Tire Dynamics}
\label{Tire Dynamics}

Tire forces are determined based on the friction curve for each tire $\left\{\begin{matrix} {^iF_{t_x}} = F(^iS_x) \\{^iF_{t_y}} = F(^iS_y) \\ \end{matrix}\right.$, where $^iS_x$ and $^iS_y$ represent the longitudinal and lateral slips of the $i$-th tire, respectively. The friction curve is approximated using a two-piece spline, defined as $F(S) = \left\{\begin{matrix} f_0(S); \;\; S_0 \leq S < S_e \\ f_1(S); \;\; S_e \leq S < S_a \\ \end{matrix}\right.$, with $f_k(S) = a_k*S^3+b_k*S^2+c_k*S+d_k$ as a cubic polynomial function. The first segment of the spline ranges from zero $(S_0,F_0)$ to an extremum point $(S_e,F_e)$, while the second segment ranges from the extremum point $(S_e, F_e)$ to an asymptote point $(S_a, F_a)$. Tire slip is influenced by factors including tire stiffness $^iC_\alpha$, steering angle $\delta$, wheel speeds $^i\omega$, suspension forces $^iF_s$, and rigid-body momentum ${^iP}={^iM}*{^iv}$. The longitudinal slip $^iS_x$ of $i$-th tire is calculated by comparing the longitudinal components of its surface velocity $v_x$ (i.e., the longitudinal linear velocity of the vehicle) with its angular velocity $^i\omega$: ${^iS_x} = \frac{{^ir}*{^i\omega}-v_x}{v_x}$. The lateral slip $^iS_y$ depends on the tire's slip angle $\alpha$ and is determined by comparing the longitudinal $v_x$ (forward velocity) and lateral $v_y$ (side-slip velocity) components of the vehicle's linear velocity: ${^iS_y} = \tan(\alpha) = \frac{v_y}{\left| v_x \right|}$.

\subsubsection{Aerodynamics}
\label{Aerodynamics}

The vehicle digital twin is modeled to simulate variable air drag $F_\text{aero}$ acting on it, which is computed based on the vehicle’s operating condition:
$
F_{\text{aero}} = \begin{cases}
F_{d_\text{max}} & \text{if } v \geq v_{\text{max}} \\
F_{d_\text{idle}} & \text{if } \tau_{\text{out}} = 0 \\
F_{d_\text{rev}} & \text{if } (v \geq v_{\text{rev}}) \land (G_\# = -1) \land (RPM_{w} < 0) \\
F_{d_\text{idle}} & \text{otherwise}
\end{cases}
$
where, $v$ is the vehicle velocity, $v_\text{max}$ is the vehicle's designated top-speed, $v_\text{rev}$ is the vehicle's designated maximum reverse velocity, $G_\#$ is the operating gear, and $RPM_w$ is the average wheel RPM. Apart from this, a linear angular $T_d$ drag acts on the vehicle, which is directly proportional to its angular $\omega$ velocity. Finally, the downforce acting on the vehicle is also modeled proportional to its velocity: $F_\text{down}=K_\text{down}*|v|$, where $K_\text{down}$ is the downforce coefficient.

\subsection{Sensors}
\label{Sensors}

The simulated vehicle can be equipped with physically accurate interoceptive and exteroceptive sensing modalities. The modeling and simulation aspects of these perception modalities are discussed in the following sections.

\subsubsection{Actuator Feedbacks}
\label{Actuator Feedbacks}

Throttle ($\tau$), steering ($\delta$), brake ($\chi$) and handbrake ($\xi$) sensors are simulated using a simple feedback loop. These variables keep track of the commands relayed to the vehicle actuators using a \texttt{get()} method.

\subsubsection{Incremental Encoders}
\label{Incremental Encoders}

Simulated incremental encoders measure wheel rotations $^iN_{\text{ticks}} = {^iPPR} * {^iCGR} * {^iN_{\text{rev}}}$, where $^iN_{\text{ticks}}$ represents the measured ticks, $^iPPR$ is the encoder resolution (pulses per revolution), $^iCGR$ is the cumulative gear ratio, and $^iN_{\text{rev}}$ represents the wheel revolutions.

\subsubsection{Inertial Navigation Systems}
\label{Inertial Navigation Systems}

Positioning systems and inertial measurement units (IMU) are simulated based on temporally coherent rigid-body transform updates of the vehicle $\{v\}$ with respect to the world $\{w\}$: ${^w\mathbf{T}_v} = \left[\begin{array}{c | c} \mathbf{R}_{3 \times 3} & \mathbf{t}_{3 \times 1} \\ \hline \mathbf{0}_{1 \times 3} & 1 \end{array}\right] \in SE(3)$. The positioning systems provide 3-DOF positional coordinates $\{x,y,z\}$ of the vehicle, while the IMU supplies linear accelerations $\{a_x,a_y,a_z\}$, angular velocities $\{\omega_x,\omega_y,\omega_z\}$, and 3-DOF orientation data for the vehicle, either as Euler angles $\{\phi_x,\theta_y,\psi_z\}$ or as a quaternion $\{q_0,q_1,q_2,q_3\}$.

\subsubsection{Cameras}
\label{Cameras}

Simulated cameras are parameterized by their focal length $f$, sensor size $\{s_x, s_y\}$, target resolution, as well as the distances to the near $N$ and far $F$ clipping planes. The viewport rendering pipeline for the simulated cameras operates in three stages.

First, the camera view matrix $\mathbf{V} \in SE(3)$ is computed by obtaining the relative homogeneous transform of the camera $\{c\}$ with respect to the world $\{w\}$: $\mathbf{V} = \begin{bmatrix} r_{00} & r_{01} & r_{02} & t_{0} \\ r_{10} & r_{11} & r_{12} & t_{1} \\ r_{20} & r_{21} & r_{22} & t_{2} \\ 0 & 0 & 0 & 1 \\ \end{bmatrix}$, where $r_{ij}$ and $t_i$ denote the rotational and translational components, respectively.

Next, the camera projection matrix $\mathbf{P} \in \mathbb{R}^{4 \times 4}$ is calculated to project world coordinates into image space coordinates: $\mathbf{P} = \begin{bmatrix} \frac{2*N}{R-L} & 0 & \frac{R+L}{R-L} & 0 \\ 0 & \frac{2*N}{T-B} & \frac{T+B}{T-B} & 0 \\ 0 & 0 & -\frac{F+N}{F-N} & -\frac{2*F*N}{F-N} \\ 0 & 0 & -1 & 0 \\ \end{bmatrix}$, where $L$, $R$, $T$, and $B$ denote the left, right, top, and bottom offsets of the sensor. The camera parameters $\{f,s_x,s_y\}$ are related to the terms of the projection matrix as follows: $f = \frac{2*N}{R-L}$, $a = \frac{s_y}{s_x}$, and $\frac{f}{a} = \frac{2*N}{T-B}$. The perspective projection from the simulated camera's viewport is given as $\mathbf{C} = \mathbf{P}*\mathbf{V}*\mathbf{W}$, where $\mathbf{C} = \left [x_c\;\;y_c\;\;z_c\;\;w_c \right ]^T$ represents image space coordinates, and $\mathbf{W} = \left [x_w\;\;y_w\;\;z_w\;\;w_w \right ]^T$ represents world coordinates.

Finally, this camera projection is transformed into normalized device coordinates (NDC) by performing perspective division (i.e., dividing throughout by $w_c$), leading to a viewport projection achieved by scaling and shifting the result and then utilizing the rasterization process of the graphics API (e.g., DirectX for Windows, Metal for macOS, and Vulkan for Linux).

Additionally, a post-processing step physically simulates non-linear lens and film effects, such as lens distortion, depth of field, exposure, ambient occlusion, contact shadows, bloom, motion blur, film grain, chromatic aberration, etc.

\subsubsection{Planar LIDARs}
\label{Planar LIDARs}

2D LIDAR simulation employs iterative ray-casting \texttt{raycast}\{$^w\mathbf{T}_l$, $\vec{\mathbf{R}}$, $r_{\text{max}}$\} for each angle $\theta \in \left [ \theta_{\text{min}}:\theta_{\text{res}}:\theta_{\text{max}} \right ]$ at a specified update rate. Here, ${^w\mathbf{T}_l} = {^w\mathbf{T}_v} * {^v\mathbf{T}_l} \in SE(3)$ represents the relative transformation of the LIDAR \{$l$\} with respect to the vehicle \{$v$\} and the world \{$w$\}, $\vec{\mathbf{R}} = \left [\cos(\theta) \;\; \sin(\theta) \;\; 0 \right ]^T$ defines the direction vector of each ray-cast $R$, where $r_{\text{min}}$ and $r_{\text{max}}$ denote the minimum and maximum linear ranges, $\theta_{\text{min}}$ and $\theta_{\text{max}}$ denote the minimum and maximum angular ranges, and $\theta_{\text{res}}$ represents the angular resolution of the LIDAR, respectively. The laser scan ranges are determined by checking ray-cast hits and then applying a threshold to the minimum linear range of the LIDAR, calculated as \texttt{ranges[i]}$=\begin{cases} d_\text{hit} & \text{ if } \texttt{ray[i].hit} \text{ and } d_\text{hit} \geq r_{\text{min}} \\ \infty & \text{ otherwise} \end{cases}$, where \texttt{ray.hit} is a Boolean flag indicating whether a ray-cast hits any colliders in the scene, and $d_\text{hit}=\sqrt{(x_{\text{hit}}-x_{\text{ray}})^2 + (y_{\text{hit}}-y_{\text{ray}})^2 + (z_{\text{hit}}-z_{\text{ray}})^2}$ calculates the Euclidean distance from the ray-cast source $\{x_{\text{ray}}, y_{\text{ray}}, z_{\text{ray}}\}$ to the hit point $\{x_{\text{hit}}, y_{\text{hit}}, z_{\text{hit}}\}$.

\begin{figure*}[t]
    \centering
    \includegraphics[width=\linewidth]{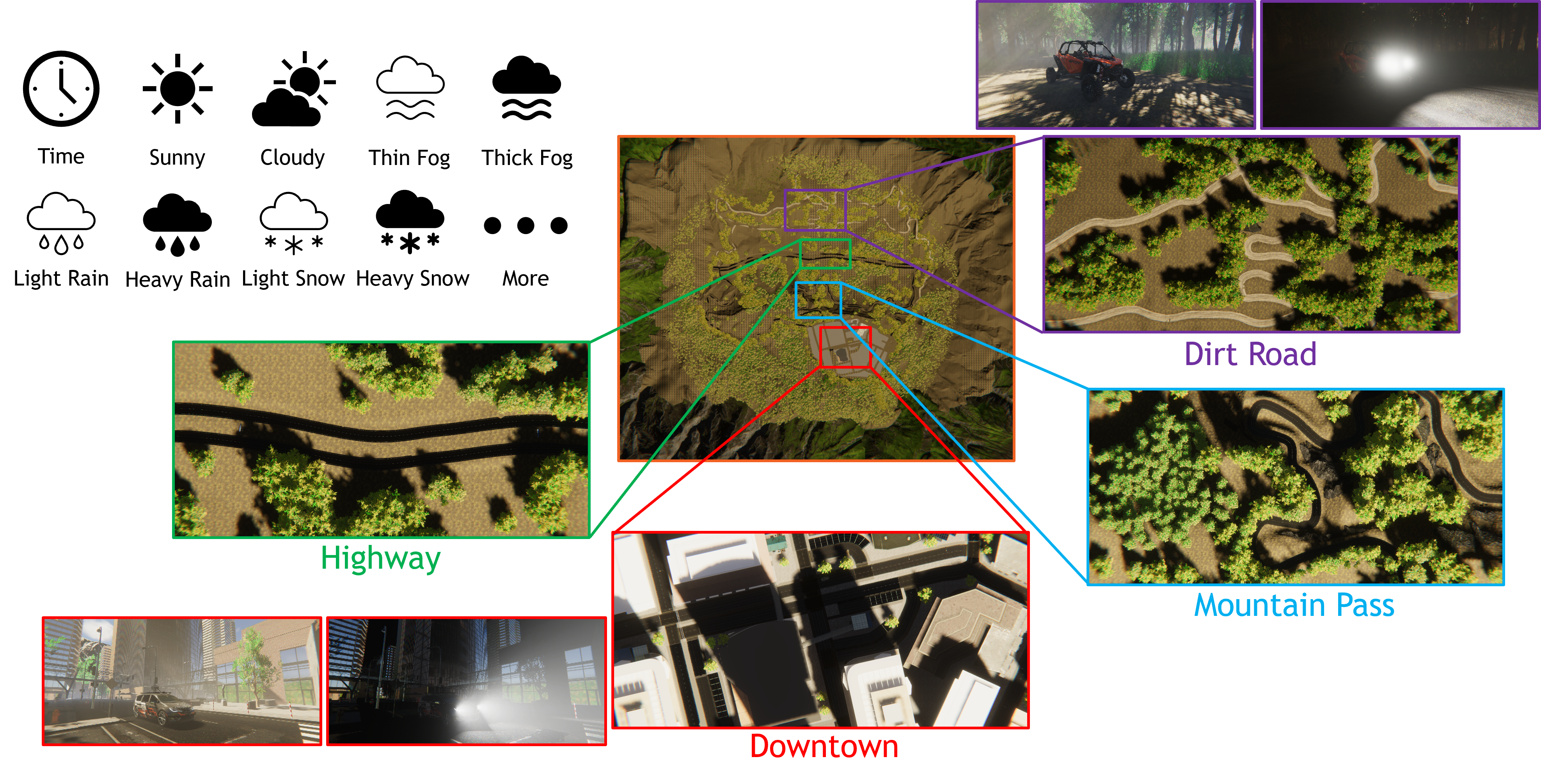}
    \caption{A feature-rich virtual environment for validating a wide spectrum of autonomy algorithms. The depicted environment is spread across 2$\times$2 km$^2$ area and comprises various driving segments such as a structured downtown, a mountain pass, a long stretch of highway, and over 4 km of dirt road. The virtual autonomous RZR was deployed in the dirt road zone.}
    \label{fig3}
\end{figure*}

\subsubsection{Spatial LIDARs}
\label{Spatial LIDARs}

3D LIDAR simulation adopts multi-channel parallel ray-casting \texttt{raycast}\{$^w\mathbf{T}_l$, $\vec{\mathbf{R}}$, $r_{\text{max}}$\} for each angle $\theta \in \left [ \theta_{\text{min}}:\theta_{\text{res}}:\theta_{\text{max}} \right ]$ and each channel $\phi \in \left [ \phi_{\text{min}}:\phi_{\text{res}}:\phi_{\text{max}} \right ]$ at a specified update rate, with GPU acceleration (if available). Here, ${^w\mathbf{T}_l} = {^w\mathbf{T}_v} * {^v\mathbf{T}_l} \in SE(3)$ represents the relative transformation of the LIDAR \{$l$\} with respect to the vehicle \{$v$\} and the world \{$w$\}, $\vec{\mathbf{R}} = \left [\cos(\theta)*\cos(\phi) \;\; \sin(\theta)*\cos(\phi) \;\; -\sin(\phi) \right ]^T$ defines the direction vector of each ray-cast $R$, where $r_{\text{min}}$ and $r_{\text{max}}$ denote the minimum and maximum linear ranges, $\theta_{\text{min}}$ and $\theta_{\text{max}}$ denote the minimum and maximum horizontal angular ranges, $\phi_{\text{min}}$ and $\phi_{\text{max}}$ denote the minimum and maximum vertical angular ranges, and $\theta_{\text{res}}$ and $\phi_{\text{res}}$ represent the horizontal and vertical angular resolutions of the LIDAR, respectively. The thresholded ray-cast hit coordinates $\{x_{\text{hit}}, y_{\text{hit}}, z_{\text{hit}}\}$, from each of the casted rays is encoded into byte arrays based on the LIDAR parameters, and given out as the point cloud data (PCD).

\subsection{Environment}
\label{Environment}

Simulated environments (refer to Fig. \ref{fig3}) can be constructed using one of the following methods:

\begin{itemize}
    \item \textbf{AutoDRIVE IDK:} Users can create custom scenarios and maps by utilizing the modular and adaptable Infrastructure Development Kit (IDK). This kit offers the flexibility to configure terrain modules, road networks, infrastructure assets, obstruction modules, and traffic elements.
    \item \textbf{Plug-In Scenarios:} AutoDRIVE Simulator supports third-party tools, such as RoadRunner \cite{RoadRunner2021}, and open standards like OpenSCENARIO \cite{OpenSCENARIO2021} and OpenDRIVE \cite{OpenDRIVE2021}. This enables users to integrate a wide range of plugins, packages, and assets in various standard formats to create or customize driving scenarios.
    \item \textbf{Unity Terrain Integration:} AutoDRIVE Simulator is developed atop the Unity game engine \cite{Unity2021} and seamlessly facilitates scenario design and development through Unity Terrain \cite{UnityTerrain2021}. Users can define terrain meshes, textures, heightmaps, vegetation, skyboxes, wind effects, and more, allowing for the creation of both on-road and off-road scenarios.
\end{itemize}

During each time step, the simulator performs mesh-mesh interference detection and calculates contact forces, frictional forces, momentum transfer, as well as linear and angular drag exerted on all rigid bodies. These forces can be used to simulate physics-based terramechanics simulations such as granular and deformable terrains.

Additionally, the time of day and weather conditions of the environment can be physically simulated. Particularly, the simulator models physically-based sky and celestial bodies to simulate varying light intensities and directions of the sun/moon using real-time or pre-baked ray-casting. This also allows the simulation of horizon gradients, as well as reflection, refraction, diffusion, scattering and dispersion of light. Additionally, the simulator leverages volumetric effects to procedurally generate different weather conditions including static and dynamic clouds, volumetric fog and mist, precipitation in the form of rain and snow particles, as well as stochastic wind gusts. The simulated time and weather can be configured to update automatically or set on-demand at predefined presets (e.g., sunny, cloudy, thin fog, thick fog, light rain, heavy rain, light snow, heavy snow, etc.) or using custom functions that control the different elements (e.g., sun, clouds, fog, rain, snow, etc.) independently. These values are accessible over the API and naturally aid in the variability analysis.


\section{HPC Deployment Framework}
\label{Section: HPC Deployment Framework}

\begin{figure*}[t]
    \centering
    \includegraphics[width=0.98\linewidth]{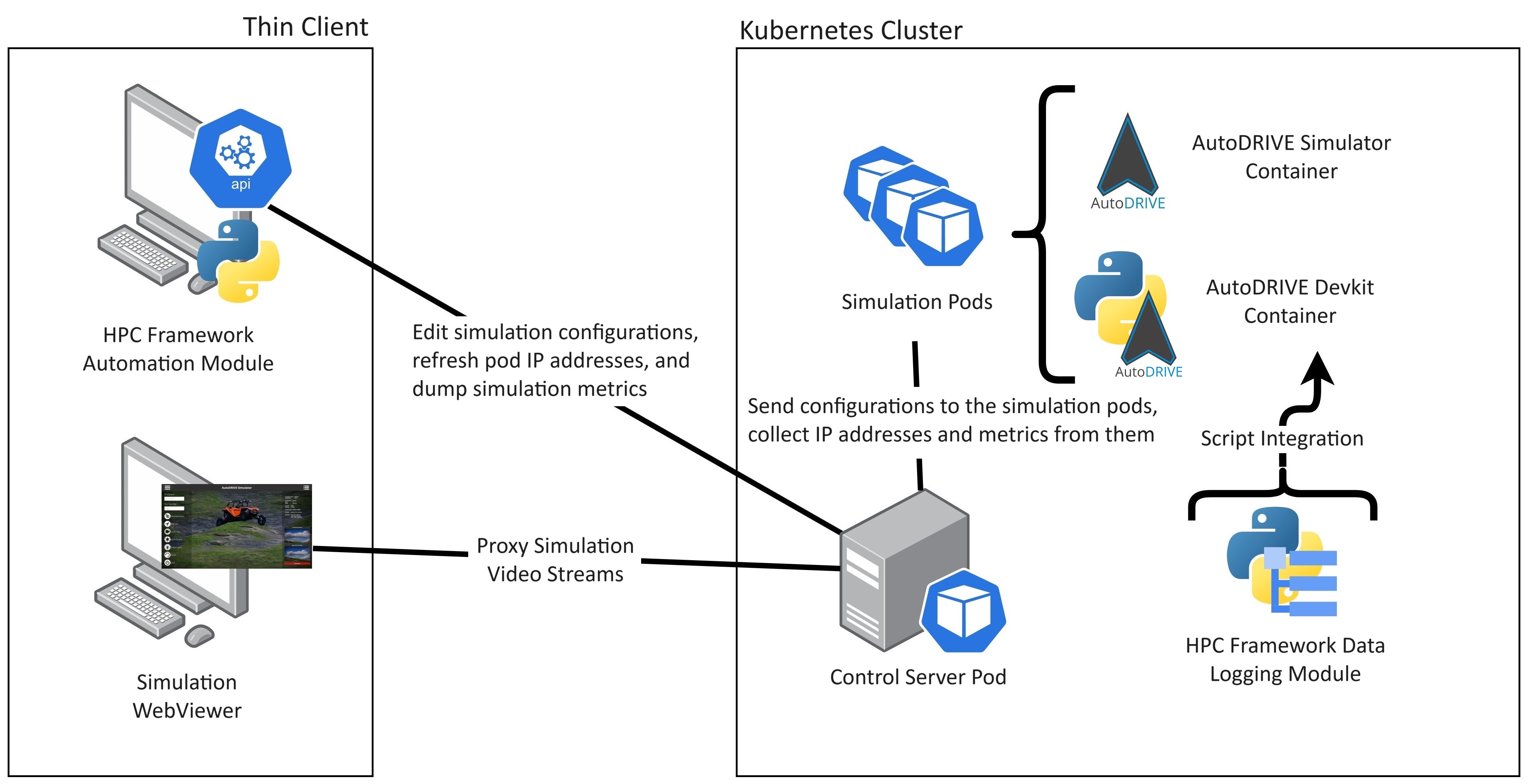}
    \caption{Overview of the HPC deployment framework for running AutoDRIVE Simulator in the cloud.}
    \label{fig4}
\end{figure*}

This section delineates key components of the cloud simulation pipeline (see Fig. \ref{fig4}). It covers essential details including the configuration of cluster hardware, the containerization process, Kubernetes configurations and specifications, Python APIs and libraries, and the functionality of the control server. 

\subsection{HPC Cluster Specifications}
\label{Section: HPC Cluster Specifications}

This work accessed HPC resources through a Kubernetes cluster managed with the Rancher cluster management service. Nevertheless, it is to be noted that all configurations discussed herein are transferable to any cluster running Kubernetes with GPU-enabled nodes. We specifically adopted Rancher to facilitate the orchestration of HPC resources within the VIPR-GS research group and work in collaboration with the Clemson Computing and Information Technology (CCIT) team to field the service before broader use.

\begin{table*}[t]
\centering
\caption{Hardware specifications of nodes in the Kubernetes cluster.}
\begin{tabular}{llll}
\toprule
\textbf{Node} & \textbf{CPU} & \textbf{Memory} & \textbf{GPU} \\
\midrule
001 & 2 $\times$ Intel Xeon Gold 6248 & 384 GB & 2 $\times$ NVIDIA Tesla V100 32 GB for PCIe \\
002 & 2 $\times$ Intel Xeon Gold 6248 & 384 GB & None \\
003 & 2 $\times$ Intel Xeon Platinum 8358 & 256 GB & 2 $\times$ NVIDIA A100 80 GB for PCIe \\
\bottomrule
\end{tabular}
\label{tab1}
\end{table*}

The sandbox cluster consists of 3 nodes, the hardware specifications of which are detailed in Table \ref{tab1}. It is worth noting that the results reported in this paper were exclusively taken on Node 003 hosting 2 $\times$ NVIDIA A100 (80 GB) cards with Kubernetes version v1.23.16+rke2r1 and NVIDIA driver version 525.125.06. The cluster hosts 2 $\times$ NVIDIA Tesla V100 (32 GB) cards as well, but they were not time-sliced and could therefore skew the results if some simulations had unrestricted access to the V100s while others were sharing the A100s.

\subsection{Containerization}
\label{Section: Containerization}

All applications used in the cluster are required to be containerized in order for Kubernetes to be able to elastically orchestrate them across compute resources. Our framework employed the Docker Engine to containerize different applications in the cluster within 4 distinct containers, which are outlined in the following sections (a sample Dockerfile is shown in Listing \ref{Listing: Dockerfile}). The Docker images were stored in a private container registry on the same network as the Kubernetes cluster.

\subsubsection{AutoDRIVE Simulator Containerization}
\label{Section: AutoDRIVE Simulator Containerization}

The AutoDRIVE Simulator container is built on top of \texttt{nvidia/vulkan} base container and installs the necessary software dependencies, an X virtual framebuffer (Xvfb) server, fast forward moving picture experts group (FFmpeg) \cite{tomar2006converting}, and a Python-based HTTP server. The Xvfb server is initialized in order to render simulation streams from a headless container. FFmpeg is used to grab frames of the virtual display and forward them to an HTTP live streaming (HLS) client, which is published to an HTTP server that is ported outside of the container. FFmpeg can also be complemented or replaced with a virtual network computing (VNC) server if GUI control is needed to interact with the simulations. Additionally, in case the simulator supports a ``render off-screen'' mode, which publishes the sensor data and any video feeds to programmable outputs, initializing an Xvfb server is not necessary.

\begin{lstlisting}[caption={Example Dockerfile}, label=Listing: Dockerfile]
FROM nvidia/vulkan:1.1.121-cuda-10.1--ubuntu18.04
ENV DEBIAN_FRONTEND=noninteractive
ENV XDG_RUNTIME_DIR=/tmp/runtime-dir
ARG VERSION

# Add CUDA repository key and install packages
RUN apt-key adv --fetch-keys "url" \
    && apt update \
    && apt install -y --no-install-recommends \
        nano \
        vim \
        sudo \
        curl \
        unzip \
        libvulkan1 \
        libc++1 \
        libc++abi1 \
        vulkan-utils \
    && rm -rf /var/lib/apt/lists/*

RUN apt update --fix-missing \
    && apt install -y x11vnc xvfb \
        xtightvncviewer ffmpeg
RUN apt update && apt install -y python3
COPY AutoDRIVE_Simulator /home/AutoDRIVE_Simulator
COPY entrypoint.sh /home/AutoDRIVE_Simulator
COPY httpserver.py /home/AutoDRIVE_Simulator

WORKDIR /home/AutoDRIVE_Simulator
RUN chmod +x \
    /home/AutoDRIVE_Simulator/AutoDRIVE.x86_64
\end{lstlisting}

\subsubsection{AutoDRIVE Devkit Containerization}
\label{Section: AutoDRIVE Devkit Containerization}

The container for the AutoDRIVE Devkit was built from the \texttt{python:3.8.10} base image. Inside the image are necessary software dependencies, the Python script controlling and initializing the simulation with AutoDRIVE Python API, as well as the HPC framework's data logging module. The data logging module exposes an API, which a developer integrates into the simulation Python script to gather simulation parameters from the control server and record any desired vehicle metrics.

\subsubsection{Control Server Containerization}
\label{Control Server Containerization}

The control server is built from \texttt{node:14}, the official node 14 Docker image, and runs a node.js webserver to orchestrate communication between simulation pods and connections outside the cluster.

\subsubsection{WebViewer Containerization}
\label{Section: WebViewer Containerization}

The WebViewer container is built off of the \texttt{nginx:alpine} image and contains a webpage that loads HLS streams from active simulations in the cluster. The WebViewer is an HTML/Javascript webpage featuring a 4$\times$4 grid of interactive windows for rendering simulation streams (refer Fig. \ref{fig1}).

\subsection{Kubernetes Configuration}
\label{Section: Kubernetes Configuration}

The cluster was configured with two deployments to orchestrate appropriate containers, three services to expose necessary ports, and an additional configuration to enable GPU time-sliced replicas. Deployments and services are applied to the Kubernetes cluster through deployment (refer Listing \ref{Listing: Deployment}) and service (refer Listing \ref{Listing: Service}) configuration files.

\subsubsection{AutoDRIVE Deployment}
\label{Section: AutoDRIVE Deployment}

The AutoDRIVE deployment is responsible for the orchestration of simulation pods running the AutoDRIVE Simulator container and the AutoDRIVE Devkit container as outlined earlier. The Devkit container communicates with the Simulator container to initialize and control simulations (send vehicle commands, analyze sensor input, change environmental parameters, etc.) over a pod's local network. Every AutoDRIVE pod requires a GPU in order to run a simulation instance, resulting in the number of replicas in this deployment being dependent on the cluster's GPU (or GPU replicas if a GPU is either time-sliced or in multi-instance mode) availability.

\subsubsection{WebViewer Deployment}
\label{Section: WebViewer Simulation Pod}

The WebViewer deployment contains only one pod, which hosts a container for the control server and another container for the WebViewer to render live video streams from all the simulation instances as they spin-up and run.

\begin{lstlisting}[caption={Example Deployment YAML}, label=Listing: Deployment]
kind: Deployment
apiVersion: apps/v1
metadata:
  namespace: hpc-simulation-framework
  name: autodrive
  labels:
    app: autodrive
spec:
  replicas: 3
  selector:
    matchLabels:
      app: autodrive
  template:
    metadata:
      labels:
        app: autodrive
    spec:
      imagePullSecrets:
        - name: container_registry_key

      nodeName: node

      containers:
      - name: autodrive
        image: "image URL"
        imagePullPolicy: Always
        env:
        - name: DISPLAY
          value: ":20"
        - name: "XDG_RUNTIME_DIR" 
          value: "/tmp/runtime-dir"
        command:
        - ./entrypoint.sh
        ports:
        - containerPort: 8000
        resources:
          limits:
            nvidia.com/gpu: 1
      - name: autodrive-api
        image: "image URL"}
        imagePullPolicy: Always
        command: ["python"]
        args: ["AutoDRIVE_Devkit/rzr_aeb.py"]
        ports:
        - containerPort: 4567
\end{lstlisting}

\begin{lstlisting}[caption={Example Service YAML}, label=Listing: Service]
apiVersion: v1
kind: Service
metadata:
  name: autodrive-headless-service
spec:
  selector:
    app: autodrive
  ports:
    - protocol: TCP
      port: 80
      targetPort: 8000
  clusterIP: None
\end{lstlisting}

\subsubsection{Services}
\label{Section: Services}

A service is employed to expose a specific aspect of a Kubernetes cluster, typically an application or endpoint, making it accessible to connections both within the cluster and externally. Two NodePort services were used to expose the control server and WebViewer containers outside of the cluster. 

Additionally, a headless service was used to expose pods in the AutoDRIVE deployment to the control server. Unlike a NodePort service, when queried, a headless service returns a list of all corresponding resource addresses. This behavior is ideal for the control server to dynamically acquire the addresses of simulation pods and proxy video streams outside of the cluster.

\subsubsection{Time Slicing Configmap}
\label {Section: Time Slicing Configmap}

In the cluster, GPUs were configured to facilitate GPU time-slicing. This approach was adopted to generate ``replicas'' of each GPU, thereby enabling the registration of multiple GPU-enabled pods per video card within a node. The preference for time-slicing over the multiple instance graphics (MIG) mode was primarily due to the incompatibility of MIG with certain graphics libraries found on prevalent HPC video cards, such as the NVIDIA A100 \cite{NvidiaDataCenter}. As mentioned earlier, our study was conducted on Node 003 (refer Table \ref{tab1}), which was time-sliced into 8 segments per card resulting in a total of 16 GPU replicas. The choice of 8 replicas per GPU was selected as a starting point after preliminary experimentation of simulator GPU utilization on a non-sliced A100 card.

\subsection{Python APIs}
\label{Section: Python APIs}

Several widely used autonomy simulators leverage a Python-based API for scripting simulations. Our framework employs a purpose-built Python library, allowing for integration with existing simulation scripts and facilitating automation scripting with relative ease.

\subsubsection{Kubernetes Python Client}
\label{Kubernetes Python Client}

The Kubernetes official Python client wraps calls to the RESTful Kubernetes API server in order to allow for complete cluster control via Python scripts. This client is used in the HPC Framework Python Library for cluster orchestration.

\subsubsection{HPC Framework Python Library}
\label{Section: HPC Framework Python Library}

A custom Python library was developed, serving two primary purposes: (i) integrating with the existing simulation scripts to accurately log simulation metrics and (ii) providing an abstracted API through which the users can automate the verification and validation of all the necessary test cases via appropriately configured simulation instances.

The aforementioned data logging module stores specified metrics in a comma-separated value (CSV) file within the simulation pod while a simulation is running. Once a test case concludes, the module contains functionality to send collected metrics back to the control plane for transmission outside of the cluster. To introduce variability among otherwise identical simulation containers, the logging module can request initialization parameters from the control plane (e.g., weather conditions, time of day, or any other variable).

The automation module allows Python scripts to automate cluster processes and run simulations in larger batches. Functionally, this is achieved through a combination of calls to the Kubernetes Python API and requests made to the control plane within the cluster.

\subsection{Control Server}
\label{Section: Control Server}

The control server sits in the middle of the HPC framework and handles communication between the simulation pods and all other systems. This section outlines the overarching communication between the server and each of the other systems.

\subsubsection{WebViewer and Control Server Communication}
\label{Section: WebViewer and Control Server Communication}

In order to account for the dynamic IP addresses of simulation pods (since pods are created and terminated after each batch of simulations) the control server opens a proxy to running simulation pod video streams by performing a DNS lookup on the headless service configured with the AutoDRIVE deployment. This allows an endpoint URL structure of \texttt{/stream/$<$sim\#$>$/video.m3u8} to route to simulations based on position in a batch regardless of that simulation's IP address. Pod addresses are updated and proxies are rebuilt by the server after each batch of simulations is completed.

\subsubsection{Pod and Control Server Communication}
\label{Section: Pod and Control Server Communication}

Outside of querying the headless service for pod addresses, the control server mediates metrics reported from simulation pods and stores them to be extracted by the HPC Framework Python Client. Once completed, simulation pods post metrics recorded to an endpoint on the control server. The control server hosts another endpoint that the simulation pods connect to using the HPC Framework Python Client in order to fetch any variation in simulation parameters.

\subsubsection{HPC Framework Python Client and Control Server Communication}
\label{Section: HPC Framework Python Client and Control Server Communication}

The HPC Framework Python Client communicates with the control server for three purposes. Firstly, to request a dump of the simulation data stored in the server and save each batch of simulations to a CSV file locally. Secondly, to post a change in the server's stored simulation configuration parameters, which are then passed to pods when they are initializing. Thirdly, to command the control server to query the headless service and refresh the stored pod IP addresses when a new batch of simulations is launched.


\section{Results and Discussion}
\label{Section: Results and Discussion}

\begin{figure*}[t]
    \centering
    \includegraphics[width=\linewidth]{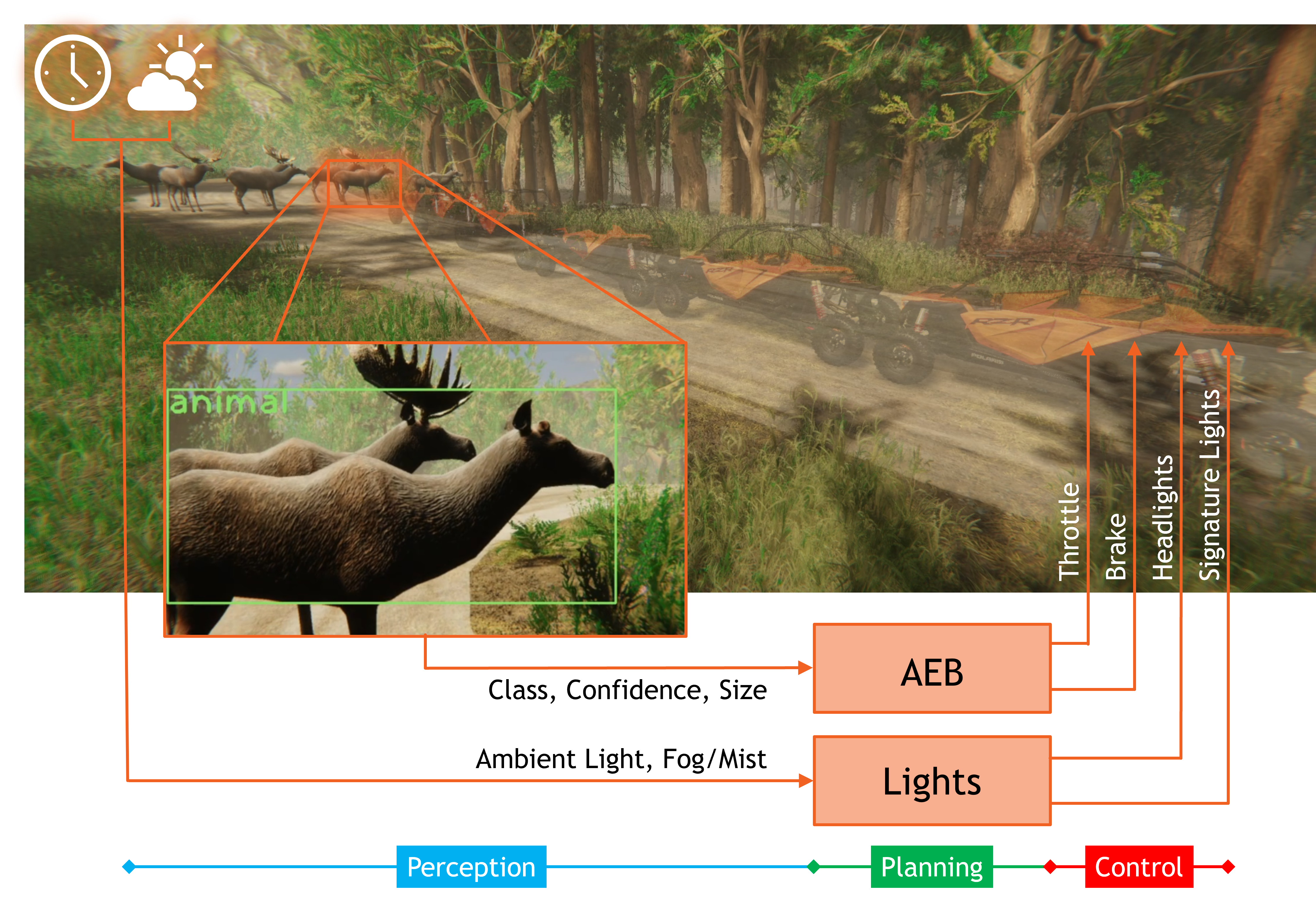}
    \caption{Candidate off-road autonomy algorithm chosen for this study in action: vision-guided autonomous emergency braking for the RZR digital twin in off-road environment.}
    \label{fig5}
\end{figure*}

For validating a candidate off-road autonomy algorithm (i.e., vision-guided autonomous emergency braking), a test scenario was ideated and implemented within AutoDRIVE Simulator (see Fig. \ref{fig5}). This test scenario dictates that an autonomous LTV (the Polaris RZR PRO R 4 ULTIMATE, in our case), which is the system under test (SUT), continues to drive straight on a dirt road while continually performing visual servoing (VS) in order to execute a panic braking maneuver in case it encounters any animal(s) along its mission path.

To this end, we devised a perception module that makes use of AI-based object detection models \cite{YOLO, YOLOv2, YOLOv3} to detect and classify objects in the environment (it is worth mentioning that these AI models are not particularly trained on data from off-road environments with objects such as moose and serve as mere candidates in this research). Furthermore, a planning strategy was formulated, which determines whether or not to trigger the autonomous emergency braking (AEB) functionality. It observes the class of the detected objects along with their sizes and classification confidence to analyze whether these objects actually exist (filter out false detections) and pose an immediate threat/liability to the vehicle (larger the object's size, more proximal it is to the vehicle). Finally, the AEB functionally controls the vehicle's throttle and brake to keep driving under nominal conditions and apply hard brakes in case a collision is imminent. Additionally, since this primary autonomy algorithm relies on visual perception, a secondary algorithm was devised to control the vehicle lights based on ambient light and fog/mist present in the environment. These values were inferred based on the time of day and weather conditions.

Taking this test scenario into account, we present the key findings and outcomes of this research in two distinct sections. Section \ref{Section: Autonomy Verification & Validation} describes the results and observations pertaining to the systematic variability analysis for validating a candidate off-road autonomy algorithm, viz. vision-guided autonomous emergency braking, to identify potential vulnerabilities in the autonomy stack's perception, planning and control modules. Section \ref{Section: Computational Analysis} then delves into the HPC resource utilization analysis under varying grades of parallel simulation workloads in order to demonstrate the effectiveness of our framework.

\subsection{Autonomy Verification \& Validation}
\label{Section: Autonomy Verification & Validation}

As described earlier, this work adopts a vision-guided AEB algorithm to illustrate the outlined V\&V workflow. It is important to acknowledge that while other perception, planning and control strategies\cite{CS4AV} can be implemented utilizing the redundant sensor suite and control degrees of freedom of the ego vehicle, the algorithm design itself is beyond the scope of this study. This work primarily focuses on the novel MSaaS framework, which is completely modular and open-source\footnote{All the project resources are made openly accessible: \url{https://github.com/AutoDRIVE-Ecosystem/AutoDRIVE-Simulator-HPC}}, thereby readily accommodating any changes in the vehicle, environment, autonomy algorithm, or the test matrix themselves.

The high-level AEB scenario was broken into multiple test cases to assess the performance of the SUT with 4 candidate perception units under test (UUT), viz. YOLOv2, YOLOv2-Tiny, YOLOv3 and YOLOv3-Tiny, across 8 different weather conditions \{clear, cloudy, thin fog, thick fog, light rain, heavy rain, light snow, heavy snow\} combined with 4 distinct times of day \{00:00 AM, 06:00 AM, 12:00 PM, 06:00 PM\}. This rather sparse test matrix quickly translated to a total of 128 test cases, which were deployed as a batch job with 8 batches, each running a set of 16 test cases, thereby covering all the 128 test conditions. It is worth mentioning that depending on the exact SUT and V\&V requirements, the total number of test cases can be devised from the resulting test matrix, and corresponding simulation instances can be spun up to carry out batched variability analysis of the algorithm.

\begin{table}[t]
\centering
\caption{High-level outcomes of the off-road autonomy verification \& validation}
\resizebox{\columnwidth}{!}{%
\begin{tabular}{llll}
\toprule
\textbf{Batch ID} & \textbf{Unit Under Test} & \textbf{Test Cases Passed} & \textbf{Total Test Cases} \\
\midrule
1 & \texttt{yolov2-tiny} & 8 & 16 \\
2 & \texttt{yolov3} & 10 & 16 \\
3 & \texttt{yolov3} & 13 & 16 \\
4 & \texttt{yolov3-tiny} & 0 & 16 \\
5 & \texttt{yolov3-tiny} & 8 & 16 \\
6 & \texttt{yolov2} & 7 & 16 \\
7 & \texttt{yolov2} & 12 & 16 \\
8 & \texttt{yolov2-tiny} & 0 & 16 \\
\bottomrule
Cumulative & N/A & 58 & 128 \\
\bottomrule
\end{tabular}%
}
\label{tab2}
\end{table}

\begin{figure*}[t]
    \centering
    \includegraphics[width=\linewidth]{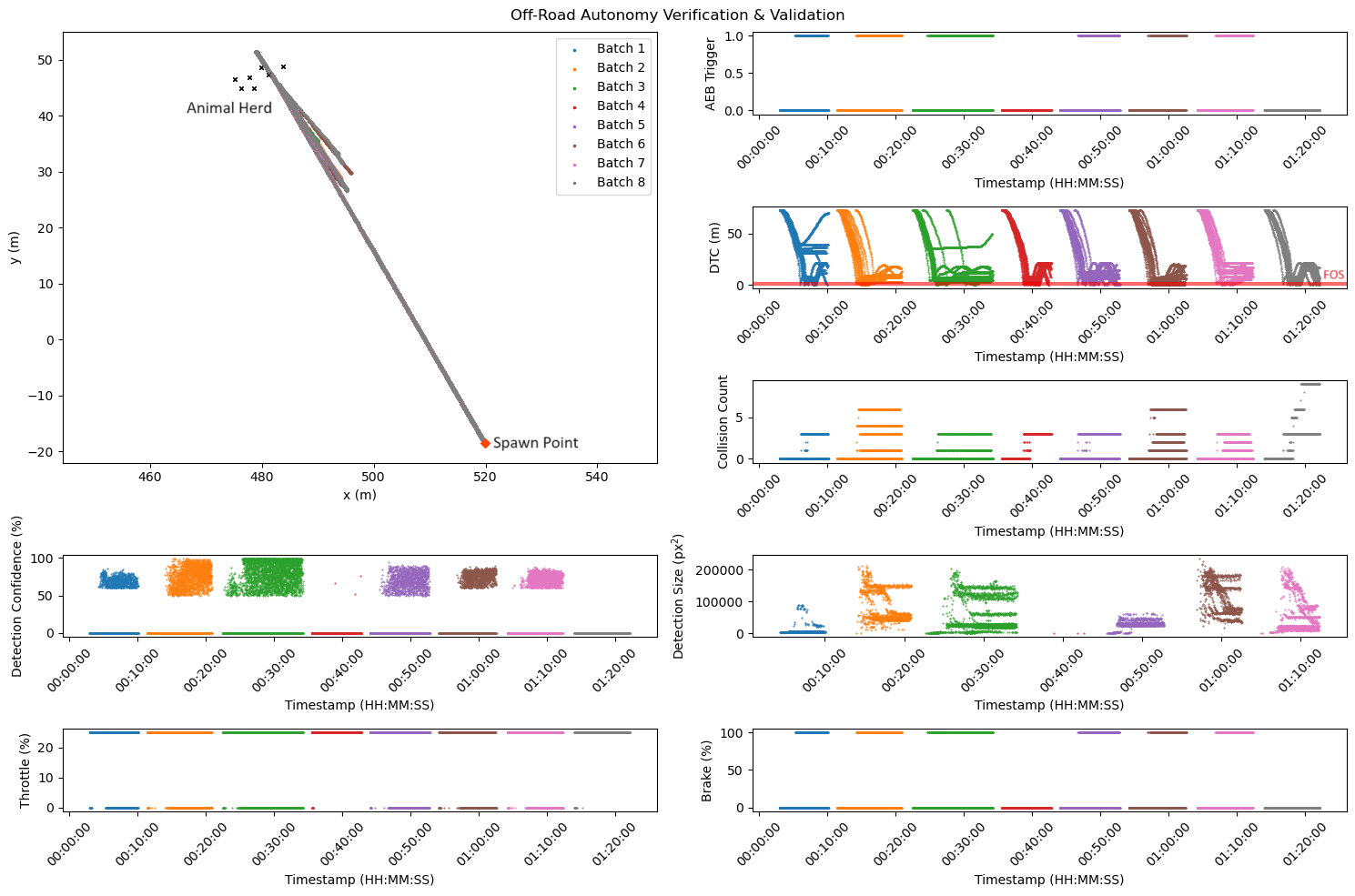}
    \caption{Detailed analysis of the candidate off-road autonomy algorithm swept across a total of 128 test cases.}
    \label{fig6}
\end{figure*}

The high-level ``pass/fail'' results of the variability analysis (refer Table \ref{tab2}) establish that YOLOv3 (71.86\% success rate) is the best candidate model for perception, while YOLOv2-Tiny (21.86\% success rate) is probably the worst, with only a small margin below YOLOv3-Tiny (25.00\% success rate). YOLOv2 performs moderately with a 59.36\% success rate. A more in-depth analysis (refer Fig. \ref{fig6}), on the other hand, provides useful insights into the potential vulnerabilities of the autonomy stack's perception, planning and control modules. This can help isolate the fault and remedy it through algorithm refinement.

Particularly, with reference to Fig. \ref{fig6}, the positional analysis reveals that the RZR would come to a safe stop in some cases, while in others, it either collided and then stopped or kept dragging along even after a collision. There are also cases where the RZR drove in reverse direction, which can be attributed to the terrain gradient, wherein, upon reducing the throttle or releasing the brakes, the vehicle freely rolled downhill due to gravitational pull. This is also evident from the distance to collision (DTC) metric, which reduces at first and then increases again.

A subsequent analysis of the AEB triggers reveals that while AEB was triggered in most cases, it was almost never triggered for test batches 4 and 8. These respectively correspond to the YOLOv3-Tiny and YOLOv2-Tiny models tested at 00:00 AM and 06:00 AM across all the weather conditions. The cause of AEB not triggering in these cases is revealed by analyzing the object detection confidence and size throughout the time series. Test batch 4 shows object detections only a handful of times with a rather average confidence, but these detections were probably quite small to trigger the AEB; these could have been long-range detections or false positives. Test batch 8, on the other hand, did not have a single detection. This naturally means that the YOLOv2-Tiny model is the worst at detecting objects. On the other hand, batch 3 (i.e. YOLOv3 at 12:00 PM and 06:00 PM across all weather conditions) has the best detections with consistent and significant confidence as well as size.

The actuator feedbacks indicate that the vehicle controller released the throttle completely when applying brakes, in order to maximize the braking performance. However, depending on several factors such as object detection consistency and carry-over velocity, the stopping distance of the vehicle sometimes exceeded the factor of safety (FOS) applied over DTC. This can be a good means of deducing the gradient of failure, which can promote rigorous testing in the optimal direction.

\begin{figure*}[t]
    \centering
    \includegraphics[width=\linewidth]{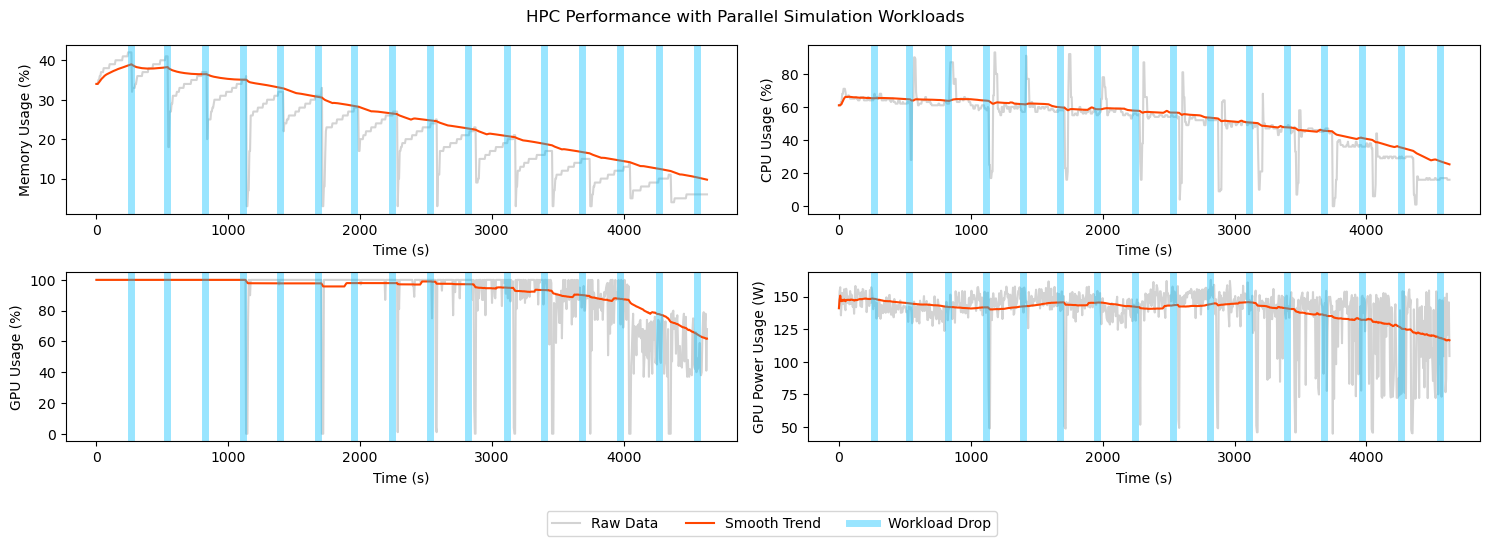}
    \caption{Performance evaluation of the HPC cluster with parallel simulation workloads. The study starts with 16 simulation instances running in parallel and 1 instance is dropped every 5 minutes, till a single instance remains.}
    \label{fig7}
\end{figure*}

The collision count decides whether a particular test case has ultimately passed or failed; a non-zero collision count implies failure. However, it can also provide a better insight if observed carefully: a higher collision count (e.g., batch 8) usually means that the vehicle collided with a moose multiple times, dragging it along with it. This typically translates to a faulty perception pipeline.

Finally, as noted earlier, a comprehensive analysis of the candidate test cases is outside the scope of this work. It is noteworthy that with the availability of logged data from various simulation instances spanning all the different test cases, a huge opportunity for data analysis opens up. The depth of analyzing this ``big data'' is primarily governed by the stakeholder requirements specifications, which vary based on the target vehicle and ODD.

\subsection{Computational Analysis}
\label{Section: Computational Analysis}

From the digital twin simulation perspective, we observed frame rates above 30 Hz while operating the simulator at its highest fidelity and upwards of 60 Hz as the simulation fidelity was reduced. Although the simulation timestep was independent of the framerate to preserve physical realism, this boost in the framerate certainly increased the real-time factor of the simulations and the overall test-case execution.

From the HPC deployment perspective, the utility of our framework is apparent from the fact that it took less than 1.33 hours to complete the entire parameter sweep of variability analysis encompassing 128 test cases. Particularly, the test cases were initialized at 12:03:05 AM and the testing continued till 01:22:10 AM. This corresponds to a total duration of 01:19:10, which is approximately equal to 1.32 hours. If the same tests were to be executed sequentially, it would take well over 10.66 hours, even without considering the transition delays. This marks a 7$\times$ reduction in testing time using the proposed framework.

To gain further insight into the computational performance aspects of our framework, we analyzed the resource utilization of the HPC cluster (refer Fig. \ref{fig7}) across different densities of parallel simulation workloads. Particularly, the evaluation commenced with 16 simulation instances operating concurrently, with one instance being killed every 5 minutes until only a single simulation instance remained. It was observed that apart from the memory consumption, which scaled roughly linearly with the workloads, the CPU and GPU usage as well as power consumption trends were rather non-linear.

This follows that running single-instance simulations sequentially is rather inefficient since (a) the computational resources available at disposal may not be utilized to their peak capacity, (b) the monetary cost incurred for increased testing time will not be compensated, and (c) the power consumption of a single-instance simulation, although lower instantaneously, is still significant and will scale several folds due to the increased testing time. Consequently, it is established that MSaaS is computationally, temporally, monetarily and ecologically more efficient and sustainable than sequential testing.


\section{Conclusion}
\label{Section: Conclusion}

This paper presented a high-fidelity digital twin simulation framework, coupled with a cloud infrastructure for elastic orchestration of simulation instances. The said framework aims at harnessing the computational power of HPC clusters to provide a scalable and efficient means of validating off-road autonomy algorithms, enabling rapid iteration and testing under a variety of conditions. We demonstrated the effectiveness of our framework through performance evaluations of the HPC cluster in terms of simulation parallelization and presented the systematic variability analysis of a candidate off-road autonomy algorithm to identify potential vulnerabilities in the autonomy stack's perception, planning and control modules. Results indicate that parallelization allowed significantly faster (7$\times$) execution of the test cases without adversely affecting the computational/monetary costs or sustainability factors.

Looking ahead, future research prospects include leveraging and enhancing the proposed framework to systematically zoom in on edge cases using a coarse-to-fine approach with active learning methods to stress-test off-road autonomy algorithms smartly. Furthermore, hardware-in-the-loop (HiL) and vehicle-in-the-loop (ViL) testing strategies can be coupled with the proposed approach to validate autonomy algorithms in realistic conditions by integrating real vehicles within simulated environments. Finally, we propose enhancing the existing MSaaS framework with an automated algorithm refinement pipeline, which could leverage the insights gained from verification and validation processes to optimally tune the parameters of the autonomy algorithm under test. These directions collectively aim to advance the reliability and performance of off-road autonomy systems across diverse environments and applications.


\bibliographystyle{IEEEtran}
\balance
{\footnotesize
\bibliography{references}
}

\end{document}